\documentclass[12pt]{article}

\usepackage{scicite}

\usepackage{times}

\usepackage{graphicx}
\usepackage{hyperref}
\usepackage{booktabs}
\usepackage{multirow}

\newenvironment{sciabstract}{%
\begin{quote} \bf}
{\end{quote}}

\title{Embodied Manipulation with Past and Future Morphologies through an Open Parametric \\Hand Design
}

\author
{Kieran Gilday$^{1\ast}$, Chapa Sirithunge$^{2}$, Fumiya Iida$^{2}$, Josie Hughes$^{1}$\\
\\
\normalsize{$^{1}$CREATE Lab, Department of Mechanical Engineering,}\\
\normalsize{Swiss Federal Institute of Technology in Lausanne, Switzerland}\\
\normalsize{$^{2}$Bio-Inspired Robotics Lab, Department of Engineering, University of Cambridge, UK}\\
\\
\normalsize{$^\ast$To whom correspondence should be addressed; E-mail:  kieran.gilday@epfl.ch.}
}

\date{}

\begin{document} 

\baselineskip24pt

\maketitle

\begin{sciabstract}
  A human-shaped robotic hand offers unparalleled versatility and fine motor skills, enabling it to perform a broad spectrum of tasks with precision, power and robustness. 
  Across the paleontological record and animal kingdom we see a wide range of alternative hand and actuation designs. Understanding the morphological design space and the resulting emergent behaviors can not only aid our understanding of dexterous manipulation and its evolution, but also assist design optimization, achieving, and eventually surpassing human capabilities. Exploration of hand embodiment has to date been limited by inaccessibility of customizable hands in the real-world, and by the reality gap in simulation of complex interactions. We introduce an open parametric design which integrates techniques for simplified customization, fabrication, and control with design features to maximize behavioral diversity. Non-linear rolling joints, anatomical tendon routing, and a low degree-of-freedom, modulating, actuation system, enable rapid production of single-piece 3D printable hands without compromising dexterous behaviors. To demonstrate this, we evaluated the design’s low-level behavior range and stability, showing variable stiffness over two orders of magnitude. Additionally, we fabricated three hand designs: human, mirrored human with two thumbs, and aye-aye hands. Manipulation tests evaluate the variation in each hand's proficiency at handling diverse objects, and demonstrate emergent behaviors unique to each design. Overall, we shed light on new possible designs for robotic hands, provide a design space to compare and contrast different hand morphologies and structures, and share a practical and open-source design for exploring embodied manipulation.
\end{sciabstract}

\section{Introduction}

The capabilities of the human hand performing robust and diverse behaviors arises through synergistic interactions between the brain and body~\cite{piazza2019century,puhlmann2022rbo,hughes2016soft}, demonstrating the fundamental principle of Embodied Intelligence (EI), that cognition is deeply rooted in the body's interactions with the world~\cite{wilson2002six, clark1998being}. 
Specifically for the hand, EI describes the synergy between the capabilities of the brain, i.e. the control of actuators or muscles, and the body, i.e. the hand morphology and passive properties of the hand. 
For a given environment or required task space animals exploit their actuation and control to achieve complex motions with a finite set of actuators, whilst the morphology and passivity provide physical robustness and behavioral diversity through local processing of physical stimuli. 
Thus, across animals we see different hand and brain structures, each of which seek to leverage actuation and hand morphology to achieve dexterous behaviors.
For example, the human hand is highly specialized for precision gripping and fine motor tasks, while the chimpanzee hand, adapted for climbing and brachiation, features longer, more curved fingers and shorter thumbs optimizing it for holding strength but limiting precision~\cite{marzke1996chimpanzee}. 
Furthermore, hand structure and functionality undergo significant changes throughout growth and development. During this period, the human hand experiences enhancement in grip strength and stability, transitioning from exploration in childhood to refined, complex use in adulthood~\cite{edwards2024hand}.
Understanding the interplay between hand morphology, passive non-linearity, and synergistic actuation~\cite{pfeifer2007self,gilday2023sensing,liu2020bioinspired} would provide a greater understanding of the contribution of embodied intelligence, and how to design these properties for given task or environment.  This would provide a step forwards towards the ultimate goal of developing a general-purpose manipulator which could outperform a human one. To better capture these links between morphology, passivity and actuation we require a design space for a human hand which can encode the vast array of possible hand and actuator designs, and corresponding technologies for rapidly evaluating their performance. 

\begin{figure}[t]
    \centering
    \includegraphics[width=0.99\columnwidth]{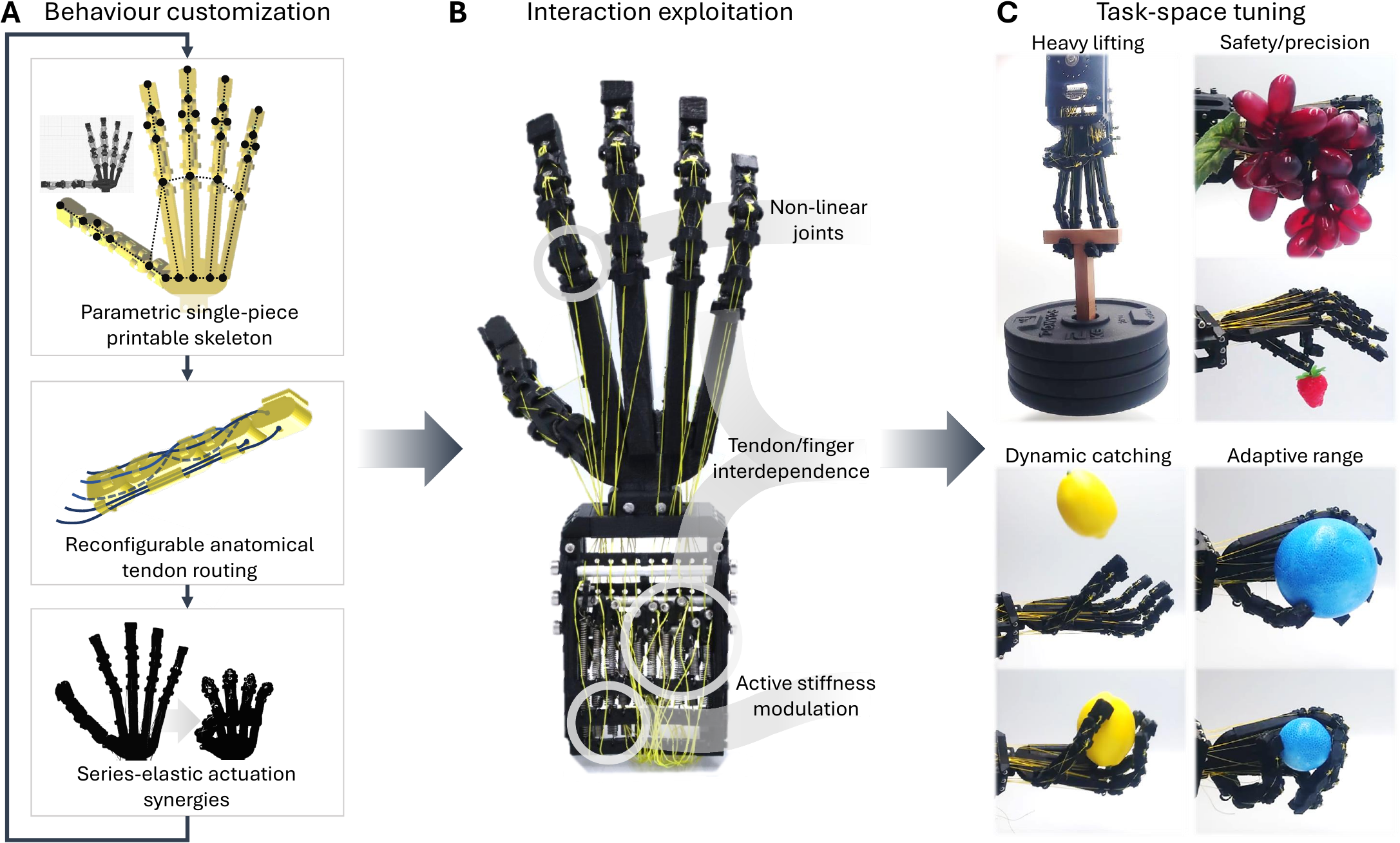}
    \caption{The open parametric design for accessible, behaviorally diverse, functional hands. (\textbf{A}) Utilising single-piece 3D printing with parametric design, anatomical tendon routing/joints, and low degree of freedom synergistic actuation hands can be rapidly customized to suit a variety of tasks. (\textbf{B}) Hand behavioral diversity expands task capabilities and emerges from complex interactions between the non-linear joints, tendon transmissions/joint coupling effects, and modulation of these by actuators. (\textbf{C}) Functionality from accessible design/manufacturing, practical uses with high strength/capabilities in a variety of tasks, and minimal actuation.}
    \label{fig:2}
\end{figure}

To develop and evaluate a wide range of hand designs, simulation is one tool which could overcome the fabrication and scaling bottleneck~\cite{akkaya2019solving,pan2021emergent}. 
However, physics simulators are still limited in their ability to capture contact forces or multi-material interactions, therefore, they fail to capture nuanced real-world interactions~\cite{bullock2011classifying,kim2021mo}. 
Thus, it remains fundamental to develop a design space for complex hands which can be fabricated and controlled in the real world, whilst maintaining their passive emergent properties.
To date, the design of dexterous robotic hands largely focuses on bio-inspired hands, which mimic the morphology of human hands to achieve similar capabilities~\cite{piazza2019century,mattar2013survey,shadow,grebenstein2012hand,puhlmann2022rbo}. 
There have been many impressive anthropomorphic hands which closely mimic the human hand bone structure and tendon routing~\cite{xu2016design,kim2019fluid,deshpande2011mechanisms,gilday2023predictive}, resulting in adaptive behaviors~\cite{xu2016design}, high performance~\cite{kim2019fluid}, and emergent grasping~\cite{gilday2023predictive}. However, these designs are often challenging to control and expensive to manufacture~\cite{deshpande2013control}, particularly so when they are fully actuated~\cite{shadow,grebenstein2012hand}. 
One approach to overcome this is to develop underactuated soft hands such as the RBO hand 3~\cite{puhlmann2022rbo}, Pisa/IIT softhand~\cite{della2018toward} or the ADAPT Hand~\cite{junge2024robust} where the compliance can provide physical robustness and compensate for lower degree of actuation and cheaper control~\cite{montufar2015theory}.
In part due to the impressive engineering of these hands, highly integrated and bespoke assemblies are complex to fabricate and redesign, therefore, their morphology, actuation or underlying passive behaviors are not readily reconfigured. 

A recent and emerging trend driven by additive manufacturing is single-print customized hands which enables rapid design test and iteration~\cite{park20243d,lee2023single,tian2018methodology,mohammadi2020practical}.
These often leverage living hinge joints such that they can be printed with a single material and require minimal assembly or post-processing~\cite{mohammadi2020practical,mannam2024design}.   
Such hands are readily customized and fabricated enabling investigation into co-design and optimization~\cite{mannam2024design}. However, these typically have poor behavioral diversity and their dexterity is limited by underactuation~\cite{mohammadi2020practical,tawk2020design}. Other designs utilize more complex rolling joints or multimaterial 3D printing to achieve greater behavioral diversity~\cite{lee2023single,bai2018self,hughes2018anthropomorphic,buchner2024replicating}. However, these designs often remain highly integrated and complex to fabricate ~\cite{park20243d,belter2013mechanical}. Additionally, these designs often over-constrain joints, reducing compliance and potentially limiting emergent behaviors and resilience~\cite{della2018toward}. Other rapidly manufacturable hand designs focus on customizability~\cite{tian2018methodology,lazaro2020graphic,bauer2022towards}, typically using parametric or modular design~\cite{mannam2024design,lazaro2020graphic,ma2013modular}.  These are largely developed for prosthetic applications, where diverse hand morphologies are necessary~\cite{mohammadi2020paediatric,kerver2023economic}. Such parametric hands use simplistic design with primitive shapes and few parameters~\cite{mannam2024design,lim2018customization,paz2017lightweight,xu2021end}. This allows some exploration of the morphological design space exploration, but does not enable exploration of underlying passive behaviors or the control and actuation~\cite{xu2016design,gilday2023sensing}.
As such, we lack a parameterized hand design that can be rapidly and accessibly fabricated, while also encoding a diverse range of behaviors ~\cite{tian2018methodology,hammond2012towards}.

We introduce the Open Parametric Hand (OPH). This 56$+$ parameter design allows diverse morphology generation through rapid redesign, single-piece 3D printing capabilities and low degree of freedom (DoF) actuation, Fig.~\ref{fig:2}A. We hypothesize that through the non-linear joints, four interconnected DoF per finger, and tendon modulation, diverse passive and active behavioral ranges can also be encoded in the morphological design space, Fig.~\ref{fig:2}B. 
To simultaneously explore control and actuation, we introduce a novel mechanism for passive-active series elastic and synergistic actuation. 
By combining these two brain and body elements we can enable the  emergence of robust and functional behaviors with a low-cost and accessible design for practical customization of hands towards new task-spaces, Fig.~\ref{fig:2}C.

\begin{figure}[t]
    \centering
    \includegraphics[width=0.99\columnwidth]{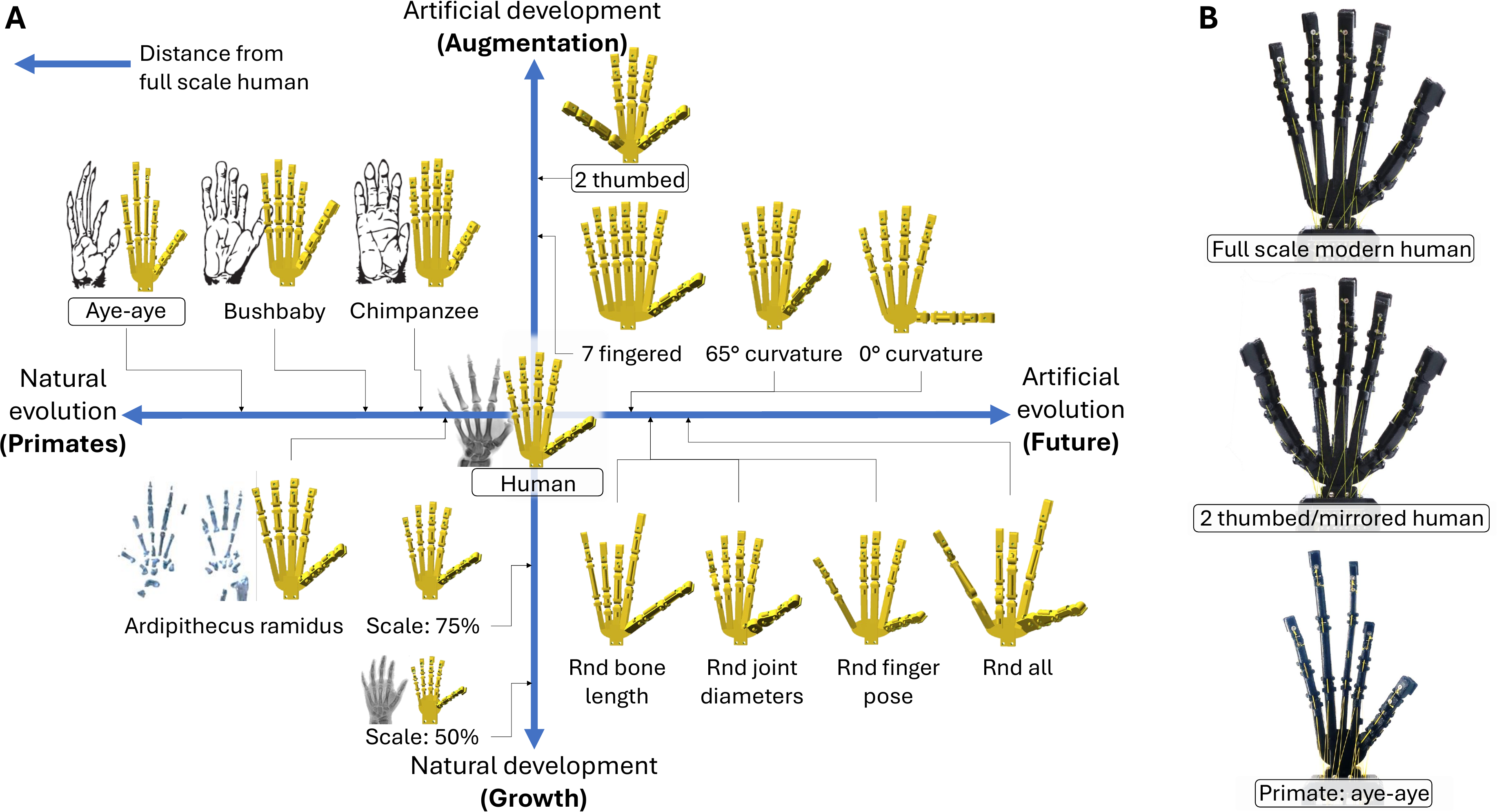}
    \caption{Morphological design space enabled by open parametric design. (\textbf{A}) Exploration of 
    embodied manipulation with short term developmental axis and long-term evolutionary axis (Primate hands from~\cite{almecija2017hands}, early hominid Ardipithecus ramidus 4.4 million years ago~\cite{white2009ardipithecus}). Distance from centre (full-scale modern human hand) measures dissimilarity. Study of `natural' hand variations can aid development of more capable `artificial' designs. (\textbf{B}) Highlighted human, aye-aye and two thumbed/mirrored hands are fabricated and prove range of design.
    }
    \label{fig:1}
\end{figure}

To structure the exploration of the OPH's customizability we propose a novel design space. This is inspired by both evolutionary and developmental processes, but has sufficient room to describe futuristic, artificially evolved hands, or expert-crafted, augmented hands, Figure~\ref{fig:1}A. We structure this as a design space with two axes, one, corresponding to long-time-scale changes in hands, considering hominid and primate hands and optimization towards environmental niches, with the potential for artificial evolution through genetic algorithms or topology optimization~\cite{howard2019evolving,xu2021end,pinskier2024diversity}. The second axis corresponds to short-time-scale development. 
On one side, this considers natural growth and on the other, artificial development, which encompasses designs continuing after full human growth through augmentation~\cite{kieliba2021robotic}, or entirely new designs through expert-lead or generative design.  The OPH can capture this design space (Fig.~\ref{fig:1}B) allowing for wide exploration.

To demonstrate this new hand design space and the OPH technologies we first characterize the novel joint and finger design, demonstrating stable function over the design space, human-like range of motion and variation in passive compliance up to 660\% through tendon reconfiguration and 400\% through parametric design. Then, to further exploit the passive behaviors, series-elastic, synergistic actuation is considered, enabling functional manipulation skills, including diverse grasp types (power and precision~\cite{feix2015grasp}), dynamic behaviors, and load capabilities up to 80~N. Finally, we showcase three representative designs (Fig.~\ref{fig:1}B) corresponding to a human hand, a mirrored hand with two thumbs, and an aye-aye (a long-fingered primate) hand. With these we highlight the unique behaviors enabled by changing morphology and interaction behaviors, such as weight-lifting and writing with the human design, multi-object grasping with the mirrored hand and arboreal locomotion and foraging with the aye-aye hand. The ability to display unique manipulation capabilities depending on morphology can be seen as an embodiment of emergence~\cite{goldstein1999emergence}, where complex systems exhibit behaviors that are unpredictable from the sum of their parts. Overall, this parametric hand opens up exciting possibilities in designing emergent behaviors and research into evolutionary and developmental underpinnings of dexterous manipulation~\cite{napier1962evolution,fragaszy2016functions}, in addition to more applied use cases. For example, task optimized manipulation solutions~\cite{pan2021emergent,chen2021co,cordella2016literature,rosati2013fully,zhang2020state}, prosthetic devices personalised both for aesthetics and functionalities~\cite{biddiss2007consumer,saikia2016recent}, and meta-human augmentation such as an additional thumb~\cite{kieliba2021robotic}.

\section{Results}
The Open Parametric Hand (OPH) is rapid and accessible to prototype and maintains the underlying passive properties and non-linear joints seen in hands. Combined with tendon driven actuation, behavioral diversity is provided whilst reducing control complexity, Fig.~\ref{fig:2}. The hand builds upon three key technologies.

First, the parametric design relies on a non-linear, dislocatable, flexure-based, rolling joint which ensures resilience, local behavioral stability and diverse interactions across a range of parameterizations. Building upon this enables a significant range of possible designs, which offer customization of workspace, inter-finger interactions, and the governing geometries for force transmission (Fig.~\ref{fig:1}). 
The second technology is the anatomy-inspired tendon layout~\cite{gilday2021wrist}. With five tendons per four DoF finger, dexterity is not compromised~\cite{pollard2002tendon}, and the layout enhances certain behaviors such as shape adaptation with an underactuated finger and allows for other emergent behaviors depending on tendon connections and actuation~\cite{gilday2023predictive}.  
The third technology is modular, series-elastic actuation synergies. In conjunction with non-linear joint behaviors, local stability, and diverse tendon configurations, synergistic actuation patterns can generate high ranges of interaction forces. A further enhancement of this actuation is the ability to actively modulate interactions with single control inputs by varying passive series elasticity~\cite{vanderborght2013variable}. 

To demonstrate the range of possible hands we focus on three hands which span the axes of the design space (Fig.~\ref{fig:2}A): the full-scale modern human, scaled up aye-aye hand and a human hand augmented with a second thumb (Fig.~\ref{fig:1}B, \href{https://drive.google.com/file/d/1M2_Gu6fKNhVcGz8WVQBVXM7WCITTpokC/view?usp=drive_link}{Movie~S1}). These hands are chosen as they represent the different axes of exploration and highlight the stability and functionality with designs distant from the origin of a modern human. The aye-aye is a small primate whose hands are specialized for climbing and foraging for insects inside tree trunks~\cite{sterling2006adaptations}. The morphology differs significantly in bone width, length, and joint diameters, with the middle finger representing approximately the smallest possible scale of joint radius and width---equivalent to $\approx$50\% scale human hand (Fig.~\ref{fig:1}A)---with current manufacturing techniques. The design with two thumbs essentially mirrors the hand design about the middle finger and we use this to highlight the potential for meta-human manipulation behaviors and the ability to customize beyond traditional anthropomorphism~\cite{kieliba2021robotic,makin2020soft}.

\subsection{Open Parametric Hand Passive behaviors}

\subsubsection{Tractable, hierarchical design}

\begin{figure}[t!]
    \centering
    \includegraphics[width=0.99\columnwidth]{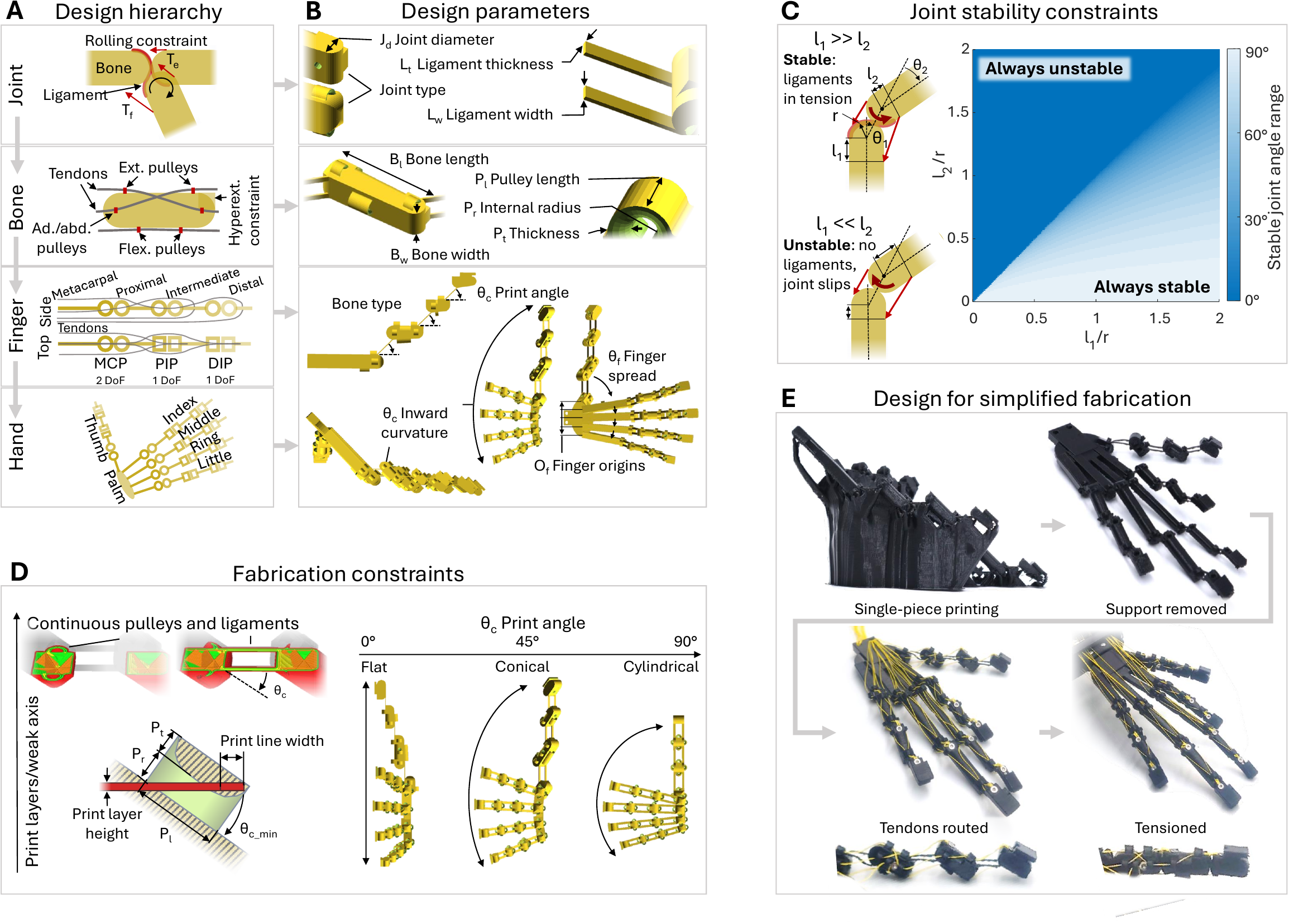}
    \caption{Parametric design for emerging complexity with stable behaviors and single-piece 3D printing capabilities. (\textbf{A}) Single ligament, non-linear, rolling contact joints assembled into bones, then fingers, then hands. (\textbf{B}) Highly customizable parameters from low-level joint parameters for stiffness/strength/range to high-level finger shapes and distributions for workspace and dexterity. (\textbf{C}) Joint design constraints. To prevent joint slipping, net tensile force should be sustained in ligament. With relative pulley placements $l_1 >> l_2$, the joint is less likely to slip (stable). (\textbf{D}) Additive manufacturing defects minimized in printing plane, thereby constraining relative finger alignment and print angle. (\textbf{E}) 3D printing example: single-piece print process with support material, tendon routing, then tendon pre-tensioning for regular joint operation.}
    \label{fig:3}
\end{figure}

The dimensionality of the OPH offers a trade-off between an intractable design space where many parameters may not influence resulting behaviors, and maintaining sufficient complexity to express a wide range of designs and maintaining the passive properties. 
To achieve this, we constrain the number of parameters using a design hierarchy and identify most critical parameters.  

At the foundation is the rolling contact joint which is partially constrained by a pair of flexure ligaments, Fig.~\ref{fig:3}A. The parameterization for the joints includes the diameter, ligament thickness, width and type. Two joint types are defined, one for the MCP joints and another for PIP/DIP joints (Fig.~\ref{fig:3}B). The MCP joint allows deflection in the abduction/adduction axis (up to 30$^\circ$) with a rounded rolling contact surface, and unlimited hyperextension. The PIP/DIP joint restricts motion in the abduction axis with cylindrical rolling contacts and includes a constraint to reduce hyperextension. Diameter, ligament thickness and width all affect joint stability, motion, friction and dislocation behaviors. 

At the next level in the design hierarchy, bones are defined as a combination of joints, tendon pulleys and anchor points for tendon terminations. Bone length and width govern the finger's workspace and strength. Pulleys constraining tendon paths are placed over each bone, and all have the same design parameters to reduce the overall complexity of the parameterization.  However, the placement of each pulley can be adjusted independently, although the relative placements and the overall distribution correlate are derived as they determine joint control and stability. Most important are the locations of the flexor tendon pulley as they critically effect joint stability. The joints are only partially constrained with a single set of ligaments, resulting in bone slip if the ligaments experience compressive forces, Fig.~\ref{fig:3}C. Pulley placement can guarantee stability across all slip conditions by controlling tendon force directions to induce tensile ligament forces. Fig.~\ref{fig:3}C visualises the joint stability range, plotting the range of angles the tendon forces will relocate bones in the most extreme slip position. 

Next in the design hierarchy is the finger design which consists of four specific bones which define the anatomical tendon layout. This layout is currently fixed with five tendons per four DoF finger, though variations could conceivably exist with different number of bones and DoFs. Finally, fingers are combined into a hand via a palm with an additional tendon path added to the thumb for enhanced palmar opposition.  Finger translation and rotation govern the hand's overall workspace. Fingers can be rotated independently in an axis normal to the palm. An inward curvature can also be defined, dependent between fingers but allowing longitudinal rotation towards the center of the palm. The palm is then derived from the finger placements. Curvature is coupled to the print angle which is a constraint placed on design for improved print quality (Fig.~\ref{fig:4}B).

Overall, Fig.~\ref{fig:3}A outlines the hand design and low-level function, with B outlining the most significant parameters which influence hand behaviors. 
The Open Passive Hand (OPH) design has been reduced down to 56 parameters (for a 4 finger, 1 thumb hand). By varying only these parameters we can already generate extreme morphological changes. Fig.~\ref{fig:1}A shows some of the possible parametric design variation from modern human hands to early hominid~\cite{white2009ardipithecus} and primate hands~\cite{almecija2017hands}, where anatomy can be mimicked from real skeletal data or extrapolated from incomplete fossil records. 

\subsubsection{Design constraints}
The hand can be fabricated through single step printing of living hinge ligaments, Fig.~\ref{fig:3}E. 
These provide high force transmission with rigid bones, low friction and resilience to impacts with compliant and dislocatable joints. The hand can be printed with widely available FDM printers using existing filaments, namely polypropylene filament, making the process low-cost and accessible. 
Layer based printing can lead to delamination, reducing part strength in the printing plane. 
This is most critical in the areas of highest stresses including the tendon pulleys and joint ligaments. 
To maximize strength, a constraint is placed on the design to ensure all ligaments and pulleys fully intersect at least one print layer, Fig.~\ref{fig:3}D. This enforces a coupling between relative finger angles which is combined into a print angle parameter with a minimum value $\theta_{c_min}$.  

Fig.~\ref{fig:3}E shows the resulting constraint and printing example of the aye-aye hand (Fig.~\ref{fig:1}B). After printing in a single piece, the tendons are fastened to their respective bone anchor points and routed through the pulleys to their actuation anchors. Once tensioned, the joints are compressed into their stable rolling configuration (\href{https://drive.google.com/file/d/1pN2JnZSr28EQqCEoRYQbPMEFNl-zc3eQ/view?usp=drive_link}{Movie~S2}).

\subsubsection{Passive finger behaviors}

\begin{figure}[t!]
    \centering
    \includegraphics[width=0.75\columnwidth]{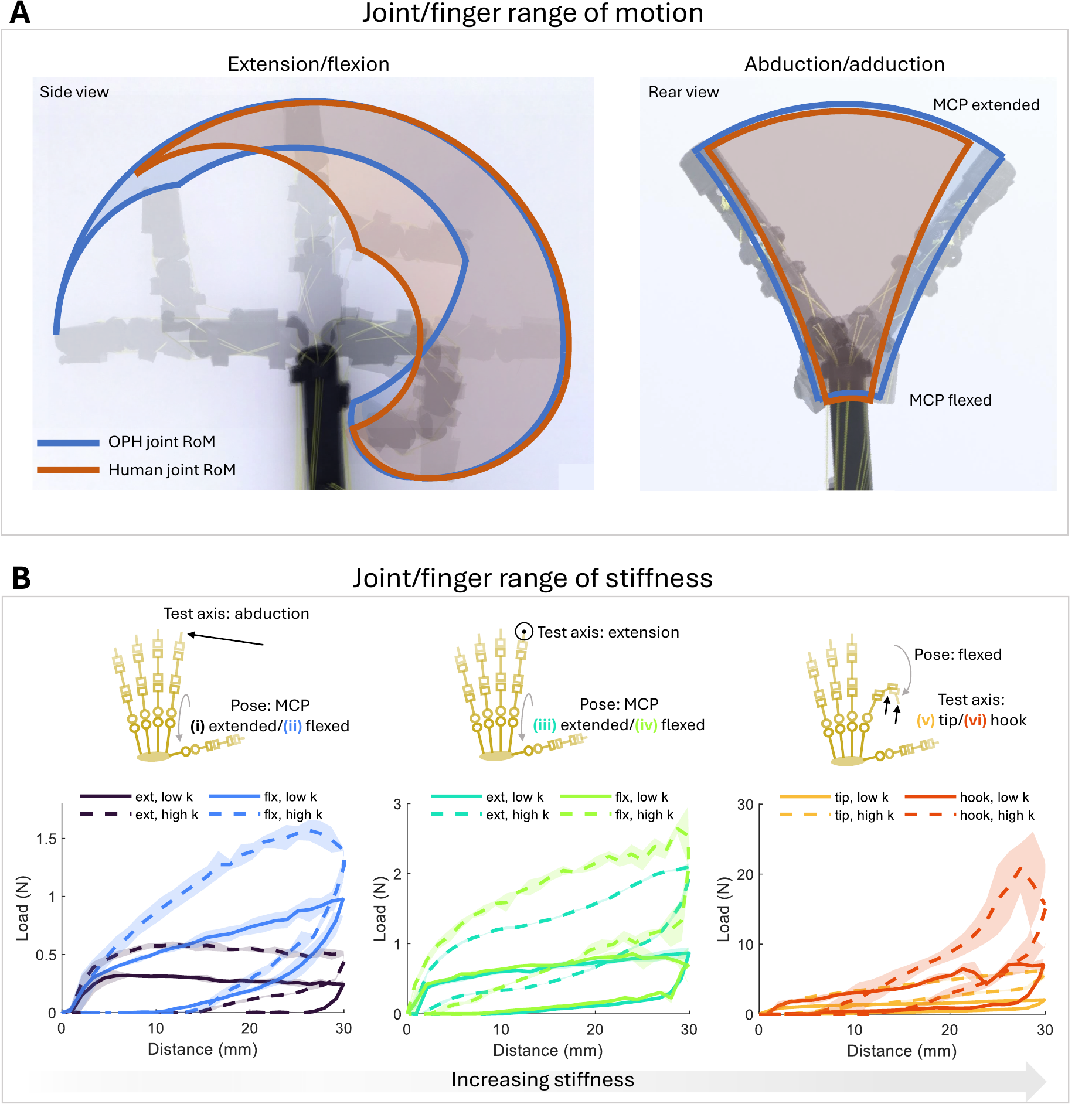}
    \caption{Joint and finger range of passive behaviors with varying tendon configuration. (\textbf{A}) joint angle and finger workspace compared to human data. (\textbf{B}) Joint/finger stiffness range with varying test axes (three sub figures), start pose, and spring stiffness (solid and dotted lines). Single variable stiffness change up to 290\%, multi-variable up to 660\%. Finger starting poses controlled by varying relative tendon lengths.
    }
    \label{fig:4}
\end{figure}

We validate the joints' range of motion (RoM) performance relative to human data. The workspace is defined by varying the tendon lengths manually to the most extreme postures shown in Fig.~\ref{fig:4}A with passive springs. 
The individual RoM of each joint is similar to a human (Table~\ref{tab:rom}) particularly so for the PIP joint and abduction axis of the MCP joint. The OPH has greater range of motion in the DIP and MCP joints, as seen in Fig.~\ref{fig:4}A. Though, overall there is reduction in the OPH workspace due to joint coupling effects. Joint coupling in the MCP and DIP can prevent simultaneous hyperextension and flexion, respectively. 
Non-linearities and joint compliance can extend the RoM. For example if the DIP is pre-flexed, the increased moment arm can prevent extension while the extensor tendon can hyperextend the MCP. External forces can dislocate joints beyond the limits found here by changing tendon lengths.

Similar to a human finger, the interacting bones, tendons, ligaments and external stimuli lead to a complex system with emergent and varied behaviors.  In particular, the emergent stiffness at different poses and in different axes highlights this, and mirrors the capabilities of a human finger.  

We measure the force/displacement of the human design index finger in different test axes, starting finger poses, and passive spring stiffness configurations, Fig.~\ref{fig:4}B (\href{https://drive.google.com/file/d/1KUWoInlx27ifjqAOR_xBnr5E9ErhAlhZ/view?usp=drive_link}{Movie~S3}). Fig.~\ref{fig:4}B left shows the stiffness characteristics in the abduction test axis. The finger is probed laterally in the extended finger pose (black) and with the MCP flexed by 90$^{\circ}$ (blue), with two different spring configurations. The first configuration is low stiffness with 0.11/0.32~N/mm springs (solid lines), the second with higher stiffness 0.32/0.87~N/mm (dashed lines). The initial extension is dominated by static friction. Likewise, the hysteresis in force when unloading is primarily from joint and tendon friction. After the initial static friction, there is an approximately linear region of stiffness and, as expected, the stiffer spring configuration generates higher finger forces in both postures. Higher stiffness is seen in the MCP flexed posture, this follows the reduced abduction RoM seen in Fig.~\ref{fig:4}A. The changing joint contact alters the relationship between tendon force and joint rotation.

In the extension axis with the finger in the extended or MCP flexed position, Fig.~\ref{fig:4}B center, the change in posture (dark to light green line) has a lesser effect than the change in spring configuration (dotted line). An increase in stiffness is observed due to the larger moment angle when the MCP is flexed. Higher stiffness is observed in this axis compared to abduction, despite the force acting in plane with all three joints, with peak forces approximately doubled. As in the abduction axis, there is significant hysteresis and an initial static friction region.

Two final stiffness tests are performed with the finger in an approximately two thirds flexed state (MCP, PIP, DIP angles: $[45^{\circ},90^{\circ},45^{\circ}]$) and probed at the fingertip (orange) in the extension axis and at the DIP (red) in the extension axis where a hook shape is formed (Fig.~\ref{fig:4}B right). The extension axis at the flexed fingertip behaves similarly to the extended and MCP flexed test, though with higher stiffness due to mechanical advantage of the probe location closer to the MCP joint. The hook test behaves uniquely, with stiffness increasing with extension. This is due to joint coupling, where extension of the MCP will cause flexion of the PIP and DIP joint, giving a caging effect on the part. %This effect is also seen in the load test, Fig.~\ref{fig:4}F, where hook grasps are effectively formed. The hook test also results in the highest peak force over 20~N.

Looking at the mean stiffness when pushing the finger in the abduction or extension axis (Fig.~\ref{fig:4}B left and center), change in finger posture gives between $\approx$0--280\% change in stiffness (extension low k, abduction low k respectively) and an average stiffness change of 130\%. With change in spring configuration in all six test cases, the stiffness can be increased between $\approx$60--220\% (abduction with MCP flexed and extension with MCP flexed respectively) (average stiffness change 160\%). By only changing test axis and controlling for posture and spring configuration, the stiffness is seen to change $\approx-$20--290\% (MCP flexed posture with low k, extended posture with high k respectively) (average $\approx$180\%).

These experiments that when varying morphology and control parameters: posture, spring configuration, and test axis, the stiffness change spans nearly two orders of magnitude. The mean measured passive stiffness can change by 660\%, from 0.0086--0.65~N/mm (extended posture/abduction axis/low k to flexed posture/hook axis/high k). This is necessary for behavioral diversity, and can be compared to the range of stiffness seen in human hands. Previous investigations have shown human finger passive stiffness in the range 0.76--2.50~N/mm~\cite{de1996two} (ellipsoid in flex plane), mimicked by variable impedance actuators~\cite{vanderborght2013variable}. Other results give stiffness 0.3--3~N/mm depending on applied forces (0.1--0.6~N/mm in just abduction)~\cite{hajian1997identification}. This is between 230\% in a single axis or 900\% when exciting the finger with different forces. While the index finger configurations we tested show lower maximum stiffness (0.65~N/mm), the stiffness ranges observed by the OPH show a similar variation (200--660\%).

\subsubsection{Design of passive finger behaviors}

\begin{figure}[t]
    \centering
    \includegraphics[width=0.99\columnwidth]{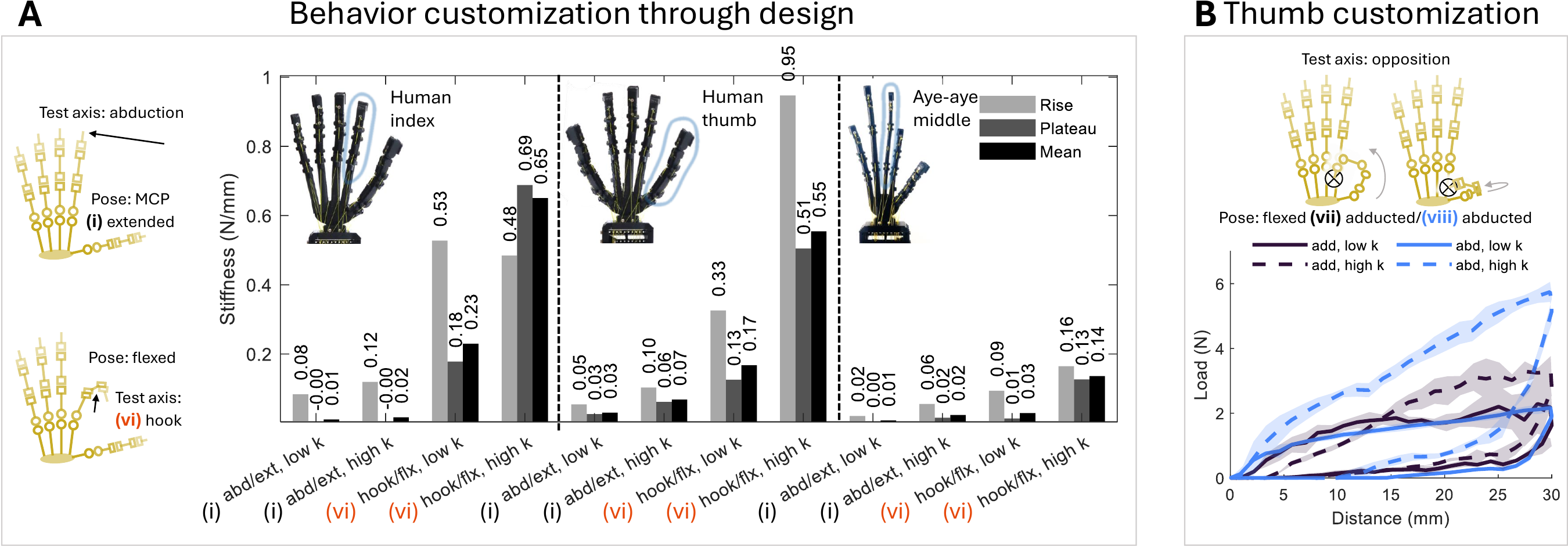}
    \caption{Joint and finger stability and range of behavior with varying parametric design.  (\textbf{A}) Stiffness range in maximum and minimum stiffness axes (Fig.~\ref{fig:4}D) with varied finger design. Up to 400\% change in stiffness by varying finger design. (\textbf{B}) Additional tendons can be added to increase thumb opposition force/stiffness.
    }
    \label{fig:5}
\end{figure}

The OPH has been used to produce three hands (human, mirrored human/two thumbed, aye-aye) by matching their skeletal geometry. Despite significant ranges of joint design, bone aspect ratios, and finger placements, all three hands have stable joints and display a range of stiffness in different poses and configurations. To demonstrate the stability and stiffness range, a single finger from each of the hands is tested. For the human hand, the index finger is used. For the mirrored hand, the thumb is chosen. For the aye-aye hand, the middle finger with the most extreme aspect ratio and smallest joints.

For the minimum and maximum stiffness poses and test axes seen in Fig.~\ref{fig:4}B, stiffness is measure with the same two passive spring configurations for each of the three fingers. The mean stiffness is calculated from the final load and displacement, the rise and plateau are calculated from lines fit to the first 5~mm and last 25~mm of displacement respectively which isolates the initial static friction phases and the spring dominated phase. % This is given for two different tendon spring configurations (low and high stiffness). 

Relative to the index finger design, the thumb is shorter by 15\% and has larger joints by 9\%, giving it mechanical advantage in transmitting force. Additionally, the thumb width is larger by 27\%, giving further mechanical advantage in the abduction/adduction axis. This is seen in Fig.~\ref{fig:5}A, where stiffness is significantly increased in the abduction axis. In the hook axis, the stiffness is not seen to change significantly. This is in part due to the larger joint diameters reducing the curling behavior of the finger, therefore the reduction in stiffness from this geometric effect opposes the increase in force transmission.

The aye-aye middle finger has the most extreme aspect ratio. The distance from the MCP to the tip is 7\% longer, while the joint diameters are reduced by 56\% and width is decreased by 51\%, greatly reducing mechanical advantage. Therefore we expect to see a significant reduction in finger stiffness. A low reduction is seen in the abduction axis, due to friction already dominating the lowest stiffness behaviors. In the hook axis, the mean stiffness is reduced by 80\%.

Thumb opposition is a significant part of human manipulation capabilities, enabling greater in-hand dexterity and stability in various power grasps. Additional tendons can be added to the thumb design to mimic intrinsic palmar muscles~\cite{bell1865hand}. For each of the three fabricated designs, an additional tendon spanning from the thumb proximal to the index metacarpal is included to increase opposition force. Thumb opposition is measured in two postures, with the thumb adducted (opposition acts approximately in the abduction axis of the thumb) and with the thumb abducted (opposition acts approximately in the extension axis). Fig.~\ref{fig:5}B illustrates these axes and the stiffness characterisation results. Compared to the abduction and extension axes of the MCP flexed index finger, stiffness is increased by a mean of 130\%. 

Overall, we confirm stiffness ranges are variable through joint diameters and bone lengths, allowing tuning at the design stage for passive interaction forces increasing up to 400\% with the designs tested here. The complex finger design results in non-linear relation between geometry and stiffness, therefore more testing is desirable to explore and potentially maximize the range of stiffness 

\subsection{Synergistic Actuation Mechanism}

The passive properties of OPH can be further extended to offer functional behaviors for active grasping and manipulation incorporating tendon actuation. To maintain simplicity and accessibility, we limit the actuation to two degrees of freedom. With synergistic actuation and unused tendons connected to passive springs, the hand can mimic the behaviors of other underactuated hands while retaining the option to reconfigure joint stiffness and actuation patterns.

\begin{figure}[t!]
    \centering
    \includegraphics[width=0.99\columnwidth]{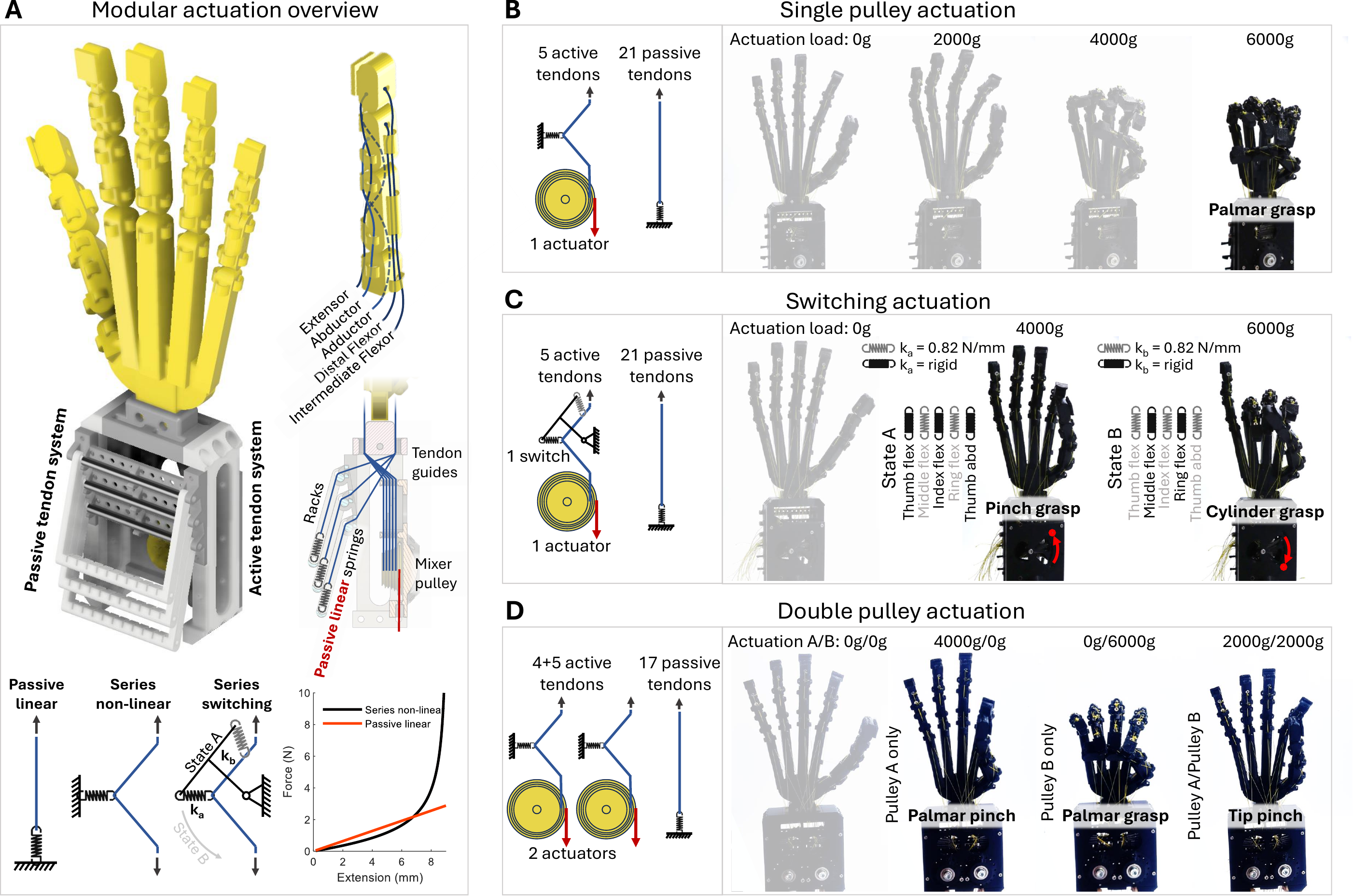}
    \caption{Low degree of actuation for posture reconfiguration and behavior modulation. (\textbf{A}) Modular actuation overview. 26+ tendons must remain in tension: active tendons with series elasticity, passive tendons with linear springs. (\textbf{B}) Single pulley actuation example: five active flexor tendons to form a palmar grasp type~\cite{feix2015grasp}. (\textbf{C}) Switching actuation example modulates series stiffness with a lever: five active tendons (thumb, index, middle and ring flexors and thumb abductor) with switching spring configuration modulate between a pinch and cylinder grasp. (\textbf{D}) Double pulley actuation example: nine active tendons (four thumb/index flexors/add/abductors on pulley A and five flexors on pulley B) for two independent grasp types (palmar pinch/palmar grasp) and interpolations (e.g. tip pinch).
    }
    \label{fig:6}
\end{figure}

Fig.~\ref{fig:6}A shows an overview of the modular actuation system which enables passive and active tendons to be routed to the hand.  Passive tendons, those providing joint compression and finger restoring forces, are routed to a compact set of racks holding linear springs. Active tendons are routed to a mixer pulley. To ensure tension, these are routed through a series, non-linear elastic system. This gives joint compliance, high force transmission when desired, and enables an enhancement where the governing spring can be switched online to obtain different behaviors between relative active tendons.  We present three configurations of the modular actuation system in Fig.~\ref{fig:6}B, C, and D which offer increasing behavioral ranges.

\subsubsection{Single Degree of Freedom Actuation}

Fig.~\ref{fig:6}B shows the single pulley actuation system. This compact design connects the distal flexor tendon of each finger to a mixer pulley with stages' diameters scaled by the relative finger joint diameters (see Fig.~\ref{fig:s2}A for schematic). The remaining tendons are connected to passive springs (see Fig.~\ref{fig:s1} for full configuration). Emerging from the base of the design is a single tendon. Thus, a single motor can actuate this synergy to enable each finger to curl from an extended state to a palmar grasp~\cite{feix2015grasp}. 

The passive behaviors can be further modified by changing the springs' stiffness and pre-tension~\cite{gilday2021wrist}. Pre-tension also allows control over starting posture and active tendon choice and mixer pulley diameters govern the final posture.

\subsubsection{Single DoF Switching Actuation}

By exploiting non-linearities in the joint moment arm, tendon friction, and series elasticity, and considering actuation as a modulation of underlying hand behaviors, a natural extension of the single pulley actuation system is one where the series spring properties can be tuned online. Fig.~\ref{fig:6}A and C show the schematic of this switchable stiffness mechanism: with a pair of springs on a binary lever for each active tendon. With switchable spring stiffness, the compliance properties during hand actuation and interactions change. 

Fig.~\ref{fig:6}C illustrates the switching stiffness mechanism. The lever holds five pairs of springs to simultaneously switch all tendons connected to the mixer pulley (see Fig.~\ref{fig:s2}B for schematic). In the example routing, state A connects the thumb distal flexor, index distal flexor and thumb abductor to rigid links (spring stiffness approximately infinite) and the middle and ring distal flexors to springs with 0.87~N/mm stiffness. State B inverts these stiffnesses. The effect is, in state A, greater actuation force is transmitted to the thumb and index finger to form a precision pinch grasp~\cite{feix2015grasp} with high stiffness. State B then relaxes the index and thumb and increases the middle and ring finger forces to form a more compliant pose resembling a power cylinder grasp. Therefore, significant behavior variation can be seen with switching the position of a binary lever while externally actuating a single tendon. 
This is especially relevant for prosthetic applications, where the ability to form precision pinches and enveloping power grasps covers a significant proportion of daily manipulation tasks~\cite{bullock2011classifying}. %This solution increases weight from 129~g to 139~g. 

\subsubsection{Two Degree of Freedom Actuation}

A second approach to extending the behavioral range it to introduce additional actuators. By having two mixer pulleys, each independently actuated, multiple actuation synergies can be generated and combined. 

Fig.~\ref{fig:6}D shows the double pulley actuation example. In this routing, each pulley actuates an independent set of tendons to generate different postures. The first pulley connects to the intermediate flexors of the thumb and index finger in addition to the thumb abductor and the sixth thumb tendon connecting the proximal thumb bone to the index metacarpal. Activating this pulley pulls together the pads of the fingertips similar to a palmar pinch grasp~\cite{feix2015grasp}. The second pulley connects to the same set of distal flexor tendons as seen in the single pulley actuation example (see Fig.~\ref{fig:s2}C for schematic). This actuation system is able to form a precision pinch grasp, a power enclosing grasp and, by actuating both degrees of actuation simultaneously, an distinct tip pinch grasp is observed.

\subsubsection{Single DoF Actuation behaviors}

\begin{figure}[t!]
    \centering
    \includegraphics[width=0.75\columnwidth]{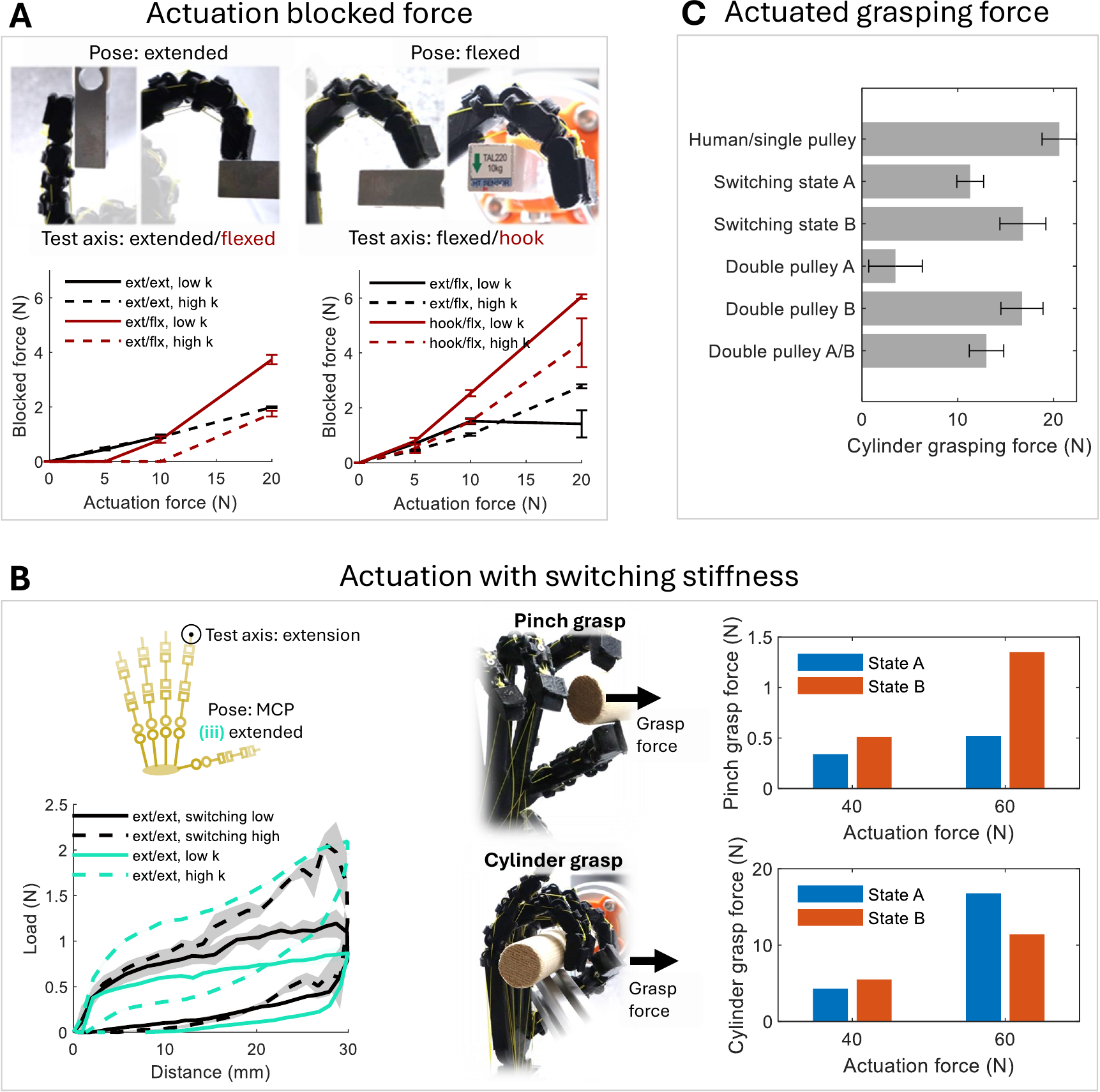}
    \caption{behavior range augmentation and modulation through actuation. (\textbf{A}) Finger force transmission range. Tendon configuration and posture affect transmission efficiency. (\textbf{B}) Switching actuation results. 80\% change in finger stiffness with online control of spring switch state, comparable to varying passive finger spring configuration (Fig.~\ref{fig:4}B). Switch states change pinch/cylinder grasping force performance to favour the state's posture. (\textbf{C}) Actuated grasping forces. Grasping force most efficient with single pulley actuation, lower efficiency with switching and double pulley especially in pinch actuation modes.
    }
    \label{fig:7}
\end{figure}

To characterise the force transmission from the actuation system we measure the blocked force when loading the distal flexor tendon (Fig.~\ref{fig:6}A) in four positions and from two starting postures and two spring configurations. Fig.~\ref{fig:7}A (left) shows the axes and results for starting in the extended position and measuring fingertip blocked force before deflection (in the extended position) and after (flexed), Fig.~\ref{fig:7}A (right) shows starting from a flexed position and measuring the force blocked at the fingertip and blocked at the PIP (hook).

Starting from the extended position, the results in Fig.~\ref{fig:7}A (left) show linear behavior once contact is made between the finger and blocking load cell. When measuring from the extended position (black), this is immediate. When measuring from the MCP flexed position (red), contact occurs at 5~N and 10~N load for the low (solid line) and high (dashed line) stiffness passive spring configuration respectively. In the stiffer spring configuration, to reach the MCP position, greater spring forces must be opposed. In the lower stiffness configuration, a greater share of the force is transmitted externally rather than to internal springs. However, for high actuation loads a fingertip curling effect is seen (MCP flexed image) which in the extended position case leads to the finger disengaging from the blocking load cell at 20~N load.

In the flexed position, Fig.~\ref{fig:7}A (right), measured blocked force at the fingertip (black) drops off at higher loads due to the finger curling and disengaging in the lower stiffness configuration (solid line). The higher stiffness passive tendons stabilize the finger and allow greater force transmission even with higher antagonistic force. The force at the PIP joint (red) shows some non-linear increase in blocked force with tendon load. This could be due to friction at lower actuation loads.

Overall, we see a trade off with joint stability and force transfer with passive spring configurations. Online adjustment of spring stiffness/tension to reduce antagonistic force when needed and increase force for greater stability otherwise, would be ideal but incurs a cost of actuation and control complexity.

\subsubsection{Switching Mechanism behaviors}

Fig.~\ref{fig:7}B shows the stiffness characterisation of the switching stiffness mechanism compared to the passive finger configuration in the extension axis (Fig.~\ref{fig:4}B center). The switching state only changes the stiffness of the distal flexor tendon compared to all five tendons of the passive configurations changing. Despite this, the switching mechanism sees an increase in stiffness of 80\% compared to 150\%.

Fig.~\ref{fig:7}B (right) shows a grasping force experiment with the two different switching states and two different grasp types measured. In the pinch grasp, a cylinder attached to a loadcell and robot arm is positioned so when the hand is actuated the index finger and thumb contact the cylinder. Grasping force is measured as the peak force when moving the cylinder with the robot arm away from the palm. In state A, the index and thumb distal flexor stiffness is greater, emphasising the pinch grasp (Fig.~\ref{fig:6}C). This is reflected in the grasping force results where, for both 40~N and 60~N actuation force, the grasping force in state A is higher.

The cylinder grasping force is measured identically to the pinch force, except the initial grasp is set to enclose the object. The results in Fig.~\ref{fig:7}B (right) show greater holding force in switching state B, where the change in spring stiffness redistributes tendon forces and all fingers contribute to a compliant power grasp. With lower actuation force, state A grasping force is higher by 1.2~N, this is due to high actuation force needed to flex all fingers into the fully enclosed state.  The switching actuation overall behaves as intended, with each state showing favourable performance in the grasp type it was designed for at high actuation forces.

\subsubsection{Comparison of Actuation Mechanism behaviors}

Fig.~\ref{fig:7}C shows the grasping force experiment repeated on the human hand design with the three example actuation systems in their different operation modes. The single pulley actuation has the highest force, followed closely by switching state B and double pulley B (Fig.~\ref{fig:6}). This is expected as both modes are designed for power grasps, and the more complex tendon routing is likely less efficient at transmitting force. The pinch grasp designed modes, switching A and double pulley A have the lowest grasping force. In particular, double pulley A actuates only the intermediate rather than distal flexor tendons so is able to transmit less force to the fingertips. Actuating both pulleys of the double pulley example simultaneously (same total tendon load 60~N) gives holding force between either mode on its own as the posture generated this way is also an interpolation.

\subsection{Emergent behaviors across the Design Space}

The hand design integrates and builds upon features at varying levels into a complex system with diverse interaction behaviors. In this section, we demonstrate the ability to vary and guide the emergence of these behaviors through customization and parameterization of the design space in Fig.~\ref{fig:1}.

\subsubsection{Grasping Performance}

\begin{figure}[t!]
    \centering
    \includegraphics[width=0.99\columnwidth]{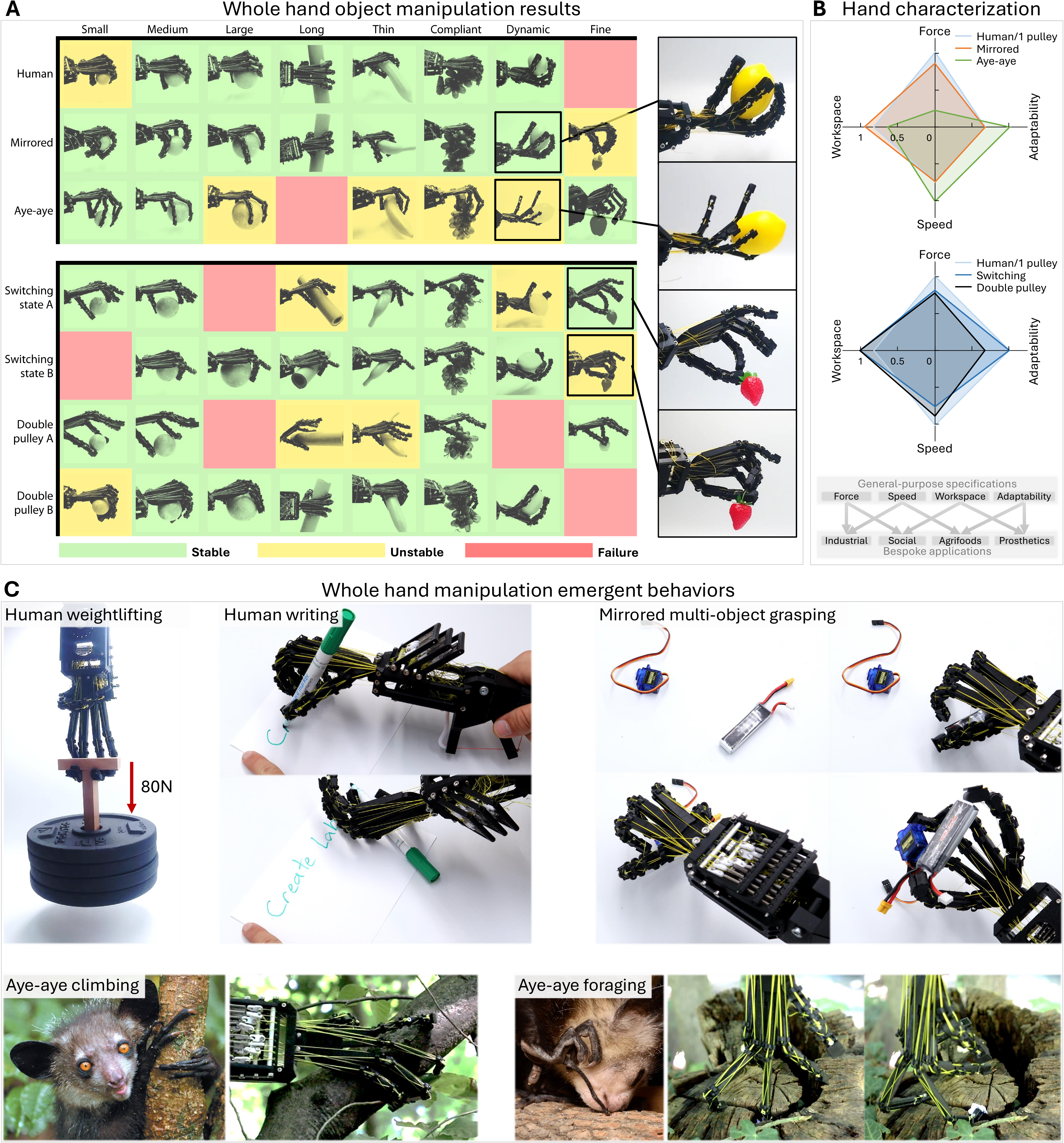}
    \caption{Emergent manipulation behaviors and hand task space tuning. (\textbf{A}) Object manipulation set for adaptive whole hand grasping. 
    (\textbf{B}) Hands and actuation systems can be biased towards different ranges of applications. 
    (\textbf{C}) Hand design unique capabilities. Weight lifting and writing with the human hand. Mirrored design multi-object grasping with sequential pinches. Aye-aye tree crawling and confined space manipulation. 
    }
    \label{fig:8}
\end{figure}

Fig.~\ref{fig:8}A shows the results of an object grasping experiment with the three hand designs (\href{https://drive.google.com/file/d/1btwGbvKrYrum2IRBlIZPH-GlDQj0cbfd/view?usp=drive_link}{Movie~S4}) and three actuation examples. The mirrored hand with two thumbs and the aye-aye hand (Fig.~\ref{fig:1}B) are actuated with the single pulley example using the mixer pulleys scaled to the relative joint diameters of each finger. All hands are controlled manually to allow exploration of diverse grasp types and the variance introduced by manual operation is compensated for by coarse result grading, either at least one \emph{stable} grasp, at best an \emph{unstable} weak grasp or the hand \emph{failing} to form any grasp.

Out of the three hand designs, the mirrored human hand has the best performance in this manipulation set. The additional abducted thumb gives greater stability in grasping objects of similar size to the hand with more enclosure options, in particular, stable grasps form readily during dynamic catching, Fig.~\ref{fig:8}A. The human hand design in the current configuration performs worse, especially unable to grasp small objects as pinching motions cannot be made. The aye-aye hand has the lowest overall performance, particularly struggling with the heavier compliant object and higher forces during dynamic catching. In these tests, grasps can often be formed, though lack the force to stably hold (compliant and thin). In others, the reduced opposability of the thumb restricts grasping (long and dynamic). However, this design has the best performance in grasping small objects such as the stalk of a strawberry (fine).

The lower half of Fig.~\ref{fig:8}A shows results of the alternative actuation systems. When allowing operation with both switching states or both of the double actuation pulleys, The human hand design is able to complete stable grasps over the whole object set. Using the switching hand in state B has similar performance when compared to the single degree of actuation hand, the cylinder power grasp compared to the palmar power grasp (Fig.~\ref{fig:6}) is able to pinch the strawberry stalk, though unable to grasp the small-sized sphere, Fig.~\ref{fig:8}A. Performance is uncompromised otherwise, and the introduction of the pinch grasp type in state A enables fine manipulation. The double pulley actuation has the same performance when comparing the second pulley actuation to the single pulley hand, expected as these form the same palmar grasp type. Pulley A covers the fine manipulation tasks to complete the overall workspace, though when compared to the switching actuation, has a worse performance in the thin cylinder and the dynamic catching tests. The two pinch grasps are distinct and the switching actuation example is more suited to this object set.

These results demonstrate the ability to customize the morphology and control to achieve different grasping capabilities. The strong performance of the double thumb hand also identifies that there are potential hand morphologies which could offer advantages over human ones.

\subsubsection{Performance Metrics}

Results from experiments in Fig.~\ref{fig:4}, Fig.~\ref{fig:5}, Fig.~\ref{fig:7} and Fig.~\ref{fig:8} are used to directly compare different aspects of the hand and actuation designs. We compare the following metrics: `workspace,' `force,' `adaptability,' and `speed' (metric definitions in Section~\ref{sec:metrics}).

The human and mirrored hand designs are rated closely, Fig.~\ref{fig:8}B, despite the observed differences in grasp types in the object set. The aye-aye hand significantly varied in force and adaptability. Likely due to the high finger compliance, but also the varied finger distribution, many distinct grasp types were seen in the object set despite a high failure rate. For example, the small, medium and large sphere are each unique, utilising different fingers to provide grasping forces. In general, we see trades offs between these metrics of workspace and adaptability and between force and speed in the design of the hand. While these metrics are not comprehensive, they can inform design choices to optimize hands for a given task set.

Fig.~\ref{fig:8}B (lower) finally shows the hand performance metrics for the different actuation systems with the human hand design. Both switching and double pulley mechanism cover the whole workspace, though have lower force output from less efficient transmission. The adaptability measured here is quantified by the maximum number of grasp types from a single actuation mode. Therefore we see the double pulley mechanism imposes greater restriction on passive adaptability than the switching mechanism. The switching mechanism is rated lowest on speed as the switch itself slows operation when switching modes. The double pulley is rated close to the single pulley, though, just below as each pulley is nearly independent (due to friction there are some sequential effects). Overall, the actuation mechanism has a lower effect on performance than design variation, though both can be used in conjunction to optimize for different tasks. For this, further investigation of design and actuation space is required.

\subsubsection{Emergent behaviors}

%Fig.~\ref{fig:8}C+ demos and highlights...

Even with only a single degree of actuation, the three hand designs exhibit diverse behavior ranges. With performance expectations from existing hand data~\cite{kieliba2021robotic,almecija2017hands} and intuition, we showcase unique manipulation demonstrations with each hand (\href{https://drive.google.com/file/d/1efEqENjNEEqSfrZt3_rJF1KHRvctMNYw/view?usp=drive_link}{Movie~S5}). 

With the human hand design, we perform a load test utilising the emergent high strength hooking behavior (Fig.~\ref{fig:4}B) to lift up to 80~N, Fig.~\ref{fig:8}C (left). Load capacity is limited by slip from the hand, without an increase actuation force, or cause damage to the actuation system where the tendon routing is not optimized for higher loads. Additionally, a writing demonstration is performed with the human hand design. Despite the lack of a frictional skin, the hand can hold a pen and write on paper with the same single pulley actuation.

The mirrored hand design exhibits a unique pinch grasping mode with the single pulley actuation. With partial actuation, an object can be pinched between the thumbs, then another pinch can be formed between the remaining partially flexed fingers and the thumbs, Fig.~\ref{fig:8}C (right).

The aye-aye hand design is unique with spindly fingers for foraging inside confined spaces and arboreal locomotion~\cite{sterling2006adaptations}. We demonstrate the ability to achieve strong grasp on branches and also to operate in confined spaces by inserting the slender index finger into the tree and retrieving an object, mimicking how aye-aye's use their fingers like this for the retrieval of insects, Fig.~\ref{fig:8}C (lower).

\section{Discussion}
Customizable hands have the potential to bridge the gap towards more general-purpose hands. With the ability to tune behaviors to a particular task-space and leveraging compliant actuation, hands can be optimized for new multi-task applications. We see this with the compliant grasping and rigid pinching behaviors exploiting non-linear compliance in the OPH with switching and double pulley actuation (Fig.~\ref{fig:7}B). We also begin to see how morphology can influence task-space with the varied capabilities of the human, mirrored and aye-aye hand designs (Fig.~\ref{fig:8}). There remains significant work in investigating the complex morphology-manipulation skill mapping, however, the evolutionary/developmental design space outlined (Fig.~\ref{fig:1}A) provides a framework grounded in real world data allowing for more objective hand evaluation. The OPH is uniquely suited to exploring this design space, through complex but tractable parameterization and rapid manufacturing capabilities.

Key to the OPH's success is the stability of low-level functionality and the diversity of emergent behaviors across varied parameterizations. We show that the non-linear, dislocatable, rolling joint has repetitive behaviors under high geometric variation and with high force capabilities. With joint range of motion comparable to that of a real human finger, dynamic stiffness variation up to seven times (660\% increase changing tendon configuration), and a further 400\% stiffness variation through varying parameterization, interactions can be effectively modulated to suit a variety of tasks. This is illustrated particularly in the high forces and varied grasp types seen in the human hand design, enabling grasping and writing with a pen, and compliance and delicate behaviors of the aye-aye design, particularly when interacting in confined spaces or adapting to different shapes (Fig.~\ref{fig:8}C). The second contribution is through the accessibility of the design. The single-piece printing capabilities with consumer grade materials and short bill of materials ($\approx$30~g filament, bolts and nylon tendons), along with the reconfigurable and simplified actuation, results in a more practical, low-cost design for large scale real world testing.

Substituting customization for dexterity does have limitations. Primarily, the task-space of each individual hand will be limited when compared to dexterous hands, meaning the OPH fills a space between single task grippers and general-purpose dexterous hands. Additionally, the current OPH design cannot reach the full manipulation space of a real hand: 80~N holding force is high for robotic hands but low in comparison to humans. Design can be varied to increase holding force through larger joints and stiffer springs (Fig.~\ref{fig:4} and Fig.~\ref{fig:5}), though this needs further exploration and actuation system optimization. Another limitation is the surface contact properties critical for fine manipulation skills and slip prevention. Introducing a parametric sensorized skin is an essential future step and should be done in a way which remains accessible and rapid, such as with silicone casting~\cite{gilday2023predictive}. These enhancements together would improve practicality for a number of real emerging applications, for example: high force ranges and robustness to uncertainties in flexible assembly, agriculture and social robotics~\cite{rosati2013fully,blanes2011technologies,zhang2020state}, not to mention low-cost, light-weight and customizable aesthetics for prosthetic applications~\cite{cordella2016literature}.

Overall, the outcomes of our research demonstrate the significant potential of customizable hands in enhancing robotic manipulation versatility and performance. Inspired by the principles of human and primate anatomy and the flexibility of 3D printing, our findings could pave the way for the creation of more adaptable, efficient, and socially accepted robotic hands, capable of fulfilling a wide range of functions across various fields.

\section{Materials and Methods}

\subsection{Design objectives}

Inspired by the varied manipulation capabilities observed across the primate taxonomy~\cite{almecija2017hands} and the desire to characterise embodiment in manipulation, we desired a platform to benchmark diverse designs. The platform should be practical and accessible, for large-scale experimentation in the real world, while allowing for complex and emergent behaviors.

Considering the extensive demands of manipulation research, which involves repeated interactions with objects and environments, the design of our skeletal design for a robotic hand prioritised robustness to endure prolonged real-world testing. Recognising the limitations of research labs in achieving the reliability of commercially produced industrial products, we optimized the design for maximum durability of components and simplified repairs. This approach not only accommodates the need for hundreds of hours of continuous experimentation without failure but also aligns with the capabilities of 3D printing and rapid prototyping technologies, making the skeletal designs of hands a resilient and practical tool for both industry and research settings.

\subsection{Functional design features}
The accessibility requirement constrains OPH design features to enable reliable fabrication with commercial 3D printers. Simplistic tendon pulleys which protrude rather than embed allow for ease of post processing, in the case of minor print defects, and ease of tendon routing. Planar printing of pulleys and ligaments gives constraints on print angles and hand curvatures, especially if using FDM printing. For other technologies, such as SLS, with improved layer adhesion this constraint can be relaxed. Smooth tendon paths with filleted surfaces improve force transmission, though can increase overhanging surface angles and associated print defects, a balance has been found for reliable FDM printing.

\subsection{Script-based parametric design}

The Open Parametric design uses the open source, script-based CAD software OpenSCAD (source files available \href{https://github.com/kg398/100_fingers/tree/release_v1.0}{here}). The file structure follows the hierarchy of Fig.~\ref{fig:3}A. Modular joint and bone designs are assembled into fingers, which are in turn assembled into hands.  There are ten critical parameters per finger: four bone lengths, three joint angles, one width, one translation and one rotation. Global parameters are defined for pulley sizes (base length, internal radius and thickness), ligaments (width and thickness) and print angle for an additional six parameters for a total of 56. This number can be reduced further, for example by deriving the three joint diameters per finger from a single value and using default values for pulley and ligament sizes as they have minimal impact on behaviors. This would bring the total down to eight per finger plus the global curvature: 41.

In current designs, hand parameters are determined manually, for example by measuring projected geometry (Fig.~\ref{fig:s3}). Pulley, ligament and curvature parameters must be estimated from image, skeleton or fossil data. 

\subsection{Modular actuation: single pulley}
The single pulley actuation example allows for up to 30 passively connected tendons and five active tendons in a single synergy. In Fig.~\ref{fig:6}B, the five active tendons are the distal flexor tendon of each finger through the series elastic mechanism with 0.87~N/mm springs (Fig.~\ref{fig:s2}A). For the thumb, middle, index, ring and little finger, the pulley radius is scaled by joint diameter: 21.0, 19.4, 18.3, 18.3, 16.0~mm, respectively. The remaining tendons are routed to the passive spring racks with stiffness 0.32~N/mm (intermediate flexor, extensor) or 0.11~N/mm (abductor, adductor). This follows the default spring configuration (Fig.~\ref{fig:s1}).

\subsection{Modular actuation: stiffness switching}
The switching stiffness follows the same design and principles as the single pulley version. Additional space is needed for the switching system which mounts a pair of spring for each of the five active tendons (Fig.~\ref{fig:6}C). Each tendon path is offset around the lever point of rotation which in one state, only one of the spring pair at a time is coupled to the tendon force/extension. In the configuration shown in Fig.~\ref{fig:6}C, the five active tendons are the distal flexor of the thumb, middle, index, and ring finger and the thumb abductor, with mixer pulley radii: 21.0, 19.4, 18.3, 18.3, 16.0~mm, respectively. In the first state,  the two thumb tendons and the index flexor are coupled through rigid `springs.' In the second state, these tendons are coupled to 0.87~N/mm springs. The middle and ring tendons have the opposite configuration. Passive tendons are configured as with the single pulley version.

\subsection{Modular actuation: double pulley}
The double pulley system is configured to operate on two independent sets of tendons. The first pulley connects the thumb and index intermediate flexors and the thumb abductor and adductor. By actuating both abductor and adductor with non-linear series springs, the joint stiffness is increased, allowing for greater opposition and pinching force. The second pulley mirrors the single pulley actuation. For both pulleys, 0.87~N/mm springs are used in series and passive tendons are configured as with the single pulley version.

\subsection{Fabrication and assembly}

The hand and actuation systems are designed for 3D printing. FDM is the most accessible though introduces higher chance of print defects compared to SLS or SLA printing. Three hands are printed for this work (Fig.~\ref{fig:1}B), each using a Prusa MK3 printer and polypropylene filament. Design parameters are extracted from skeletal diagrams and scaled to a maximum finger length of 150~mm. The hands are printed in 6--9~hours with the only post-processing being support material removal and routing of tendons through pulleys to their respective anchor points, Fig.~\ref{fig:4}B. We identify nylon or polyamide as alternative materials with desirable properties which can be printed via FDM or SLS.

The actuation systems have optimizations for reducing manufacturing time, for example: tendon clips for rapid spring attachment and tuning of tension; pivoting racks for access to assemble and tune the dense passive array; and tendons routed through rigid pulley points to reduce part count. With the exception of mixer pulley bearings, smooth rod for three low friction bearing surfaces and fasteners, the entire actuation box is 3D printed using standard materials such as PLA. The actuation system could be further optimized for strength, wear resistance and weight.

\subsection{Experimental setup}
We characterise the parametric design at three levels. First, the low level joint and finger ranges of behavior. Second, we characterise the behavioral range and stability over varying hand parameterization. Thirdly, we test the hands fully integrated with actuation to observe overall performance and emergent behaviors. 

In the first set of experiments, a single finger configuration is varied. For range of motion, tendon pretensions are adjusted manually and angle changes are measured by image data. During stiffness testing, a UR5 robot arm equipped with a loadcell probes the finger in various test axes while the finger is affixed to a table (Fig.~\ref{fig:4}D). Displacement is recorded by robot position and force is recorded directly from the calibrated loadcell. Tendons are configured with 0.11/0.32~N/mm springs in low stiffness testing and 0.32/0.87~N/mm springs for high stiffness testing.

The second set of experiments repeats the above setup with three different finger designs (Fig.~\ref{fig:5}A). An additional experiment is performed with the thumb augmented with an opposition tendon connected to a 0.32~N/mm spring (low stiffness) or 0.87~N/mm spring (high stiffness).

The third set of tests augmented via actuation measure blocked force, switching stiffness change and grasping forces similarly using a UR5 mounted loadcell (Fig.~\ref{fig:7}). Though, hands are configured with the actuation systems in Fig.~\ref{fig:6} in all tests. During blocked force, the single actuation tendon is loaded with successive weights rather than the loadcell probing. During switching stiffness, rather than tendons being reconfigured manually, just the lever is switched. During grasping, a dowel is attached to the loadcell for grip with the whole hand (Fig.~\ref{fig:7}B right).

The relative performance of the different hand designs and actuation system examples is finally explored with an object grasping test. The hands are equipped with 3D printed manual handles for motion and actuation direct from a user (Fig.~\ref{fig:8}C). Objects are chosen to represent size and shape changes with regular spheres and cylinders. Additional objects are chosen for compliant, dynamic and delicate grasping. Each test is performed attempting to pick the object from a flat surface, except for the dynamic test where the object begins in the hand, then must be thrown and caught. Evaluation of grasps is manual and has some subjectivity. Successes and failures are generally clear, though unstable grasps are more subjective based on the difficulty to form the initial grasp and low holding force which results in object slip during lift.

\subsection{Performance analysis}\label{sec:metrics}

For comparison between the hands and actuation systems, the normalized metrics (Fig.~\ref{fig:8}B) are quantified with our own definitions from the experimental results generated: `workspace' is quantified by the task performance in the object set (average score with $stable=1$, $unstable=0.5$ and $failure=0$, Fig.~\ref{fig:8}A); `force' is quantified by maximums in the grasping force experiment for the actuation examples (Fig.~\ref{fig:7}C) and range of passive stiffness for the hand designs(Fig.~\ref{fig:5}A). `Adaptability' is quantified by the number of different stable grasp types seen during the object set, where different grasp types indicate the ability for the hand to comply to object geometries or changing environments. Finally, `speed' is taken from the slowest moving finger in each design (inversely proportional to joint diameter) or heuristically with the actuation examples from the time taken to change actuation modes.

\bibliography{refs}

\bibliographystyle{Science}

\section*{Acknowledgments}
This project has received funding from the European Union’s Horizon 2020 research and innovation programme under the Marie Skłodowska-Curie grant agreement N$^\circ$~101034260.  It has also been supported by SNSF project funding.\\
The datasets generated for this study can be found in the [] repository: %\href{https://doi.org/10.5281/zenodo.12698245}{https://doi.org/10.5281/zenodo.12698245}.
\\Source files for the Open Parametric Hand can be found here: \href{https://github.com/kg398/100_fingers}{https://github.com/kg398/100\_fingers}.

\section*{Supplementary materials}
Table S1: Hand parameters and ranges.\\
Table S2: Human and OPH joint ranges of motion.\\
Fig. S1: Manual parameterization from 2D images.\\
Fig. S2: Default passive spring routing and configuration.\\
Fig. S3: Active spring configurations for the three actuation examples.\\
Movie S1: Open Parametric Hand designs generated through OpenSCAD.\\
Movie S2: Joint stabilization under tension.\\
Movie S3: Finger stiffness experiment testing procedure.\\
Movie S4: Grasping tests on diverse objects.\\
Movie S5: Unique manipulation behaviors with example hand morphologies.\\
References \cite{ishii2020quantitative}\\

\clearpage

\section*{Supplementary Materials}

\subsection*{Hand parameterization}

\renewcommand{\thetable}{S1}
\begin{table}[h]
     \caption{Hand parameterization. Design variable: name, symbol, available ranges.}
            \centering
          \scalebox{0.85}{
         \makebox[\textwidth][c]{
    \begin{tabular}{llll}
    \toprule
      Part   & Name & Symbol& Design range \\
    \midrule
      \multirow{1}{*}{Hand}  & number of fingers & $n$ & $[1,-]$\\
    \midrule
      \multirow{9}{*}{Finger geometries ($mm$)} & bone lengths (metacarpal) & $l_{meta}$ & $[d_{mcp}/2+l{p},-]$\\
        & bone length (proximal) & $l_{prox}$ & $[(d_{mcp}+d_{pip})/2+2l_{p},-]$\\
        & bone length (intermediate) & $l_{inte}$ & $[(d_{pip}+d_{dip})/2+2l_{p},-]$\\
        & bone length (distal) & $l_{dist}$ & $[d_{dip},-]$\\
        & bone width & $w$ & $[2(t_{p}+r_{p}),-]$\\
        & joint diameter (MCP) & $d_{mcp}$ & $[2(t_{p}+r_{p}),-]$\\
        & joint diameter (PIP) & $d_{pip}$ & $[2(t_{p}+r_{p}),-]$\\
        & joint diameter (DIP) & $d_{dip}$ & $[2(t_{p}+r_{p}),-]$\\
        & ligament size (width, thickness) & $[w_{lig},t_{lig}]$ & $[[0.4,w],[0.1,0.8]]$\\
        & base pulley size (length, thickness, radius) & $[l_{p},t_{p},r_{p}]$ & $[[3,-],[1.2,-],[0.5,-]]$\\
    \midrule
      \multirow{3}{*}{Finger transformations}  & origin (per finger) ($mm$) & $[x,y,z]$ & $[-,-]$\\
        & spread (palm-axis) (per finger) ($rad$) & $\theta_f$ & $[-\pi,\pi]$\\
        & inward curvature/print angle ($rad$) & $\theta_c$ & $[tan^{-1}((t_p+r_p)/l_p),\pi/2]$\\
    \bottomrule
    \end{tabular}}}
    \label{tab:params}
\end{table}

Hand parameters (Table~\ref{tab:params}) are separated into hand configuration, finger geometries and finger placements. Parameter ranges are interdependent and recorded where possible. Parameter limitations from manufacturing limits are not included, e.g., maximum bone lengths and finger placements are practically limited by 3D printer build volume.

\renewcommand{\thefigure}{S1}
\begin{figure}[ht]
    \centering
    \includegraphics[width=0.75\columnwidth]{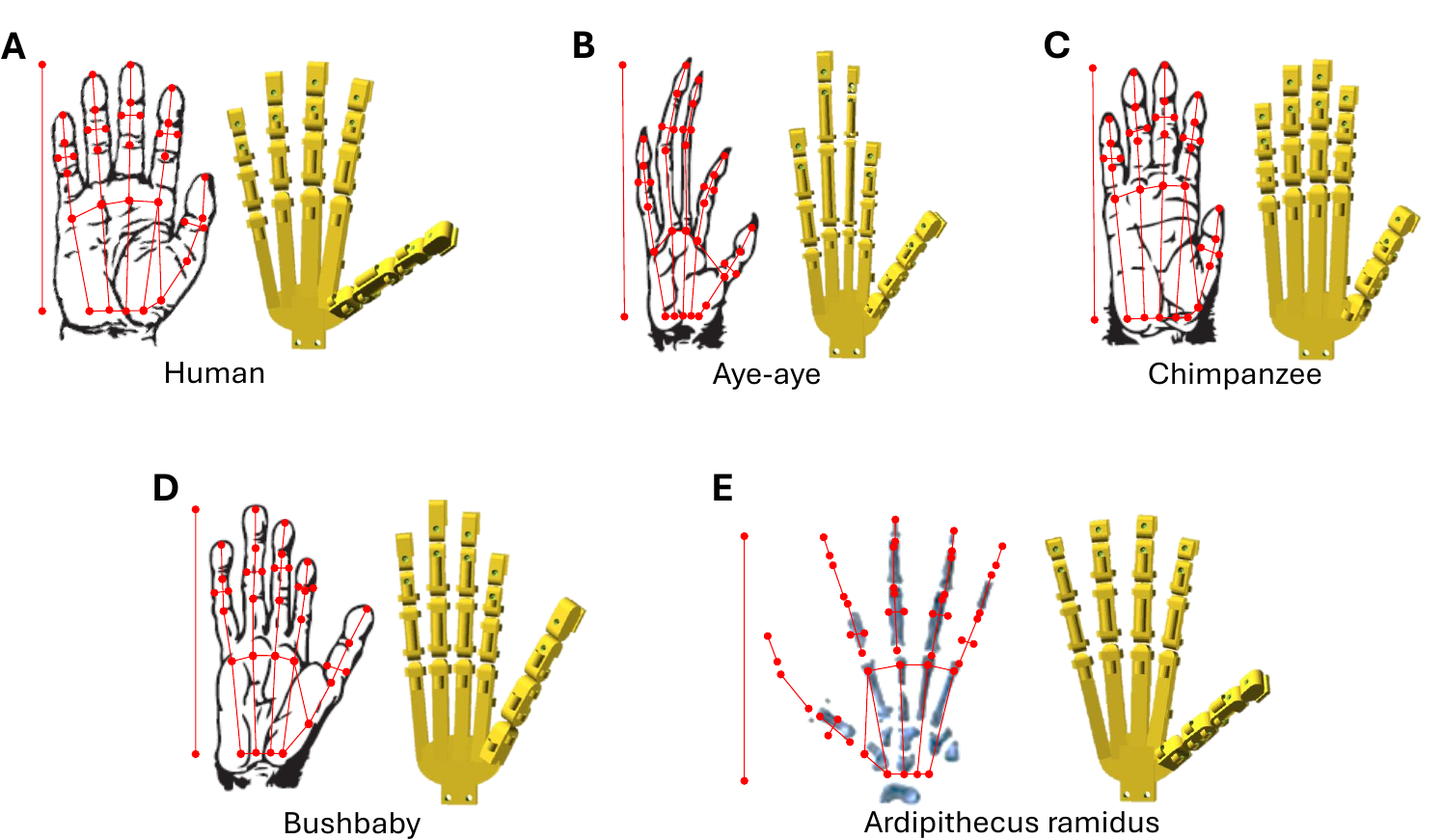}
    \caption{Manual parameter extraction from 2D hand images, sketches, or projections~\cite{almecija2017hands,white2009ardipithecus}. Approximate joint positions are identified and displacements measures. Skeletal data is measured readily or hand tracking algorithms can be used to generate initial sets of parameters.
    }
    \label{fig:s3}
\end{figure}

Basic parameterization can be generated from 2D images, Fig.~\ref{fig:s3}. Skeletal information can be inferred from existing hand tracking algorithms in particular. Some parameters cannot be gained this way, e.g., joint diameters, though will have a relationship to bone lengths and widths.

\clearpage
\subsection*{Range of motion}

\renewcommand{\thetable}{S2}
\begin{table}[h]
\caption{Comparison of human and OPH joint RoM.}
    \centering
    \scalebox{0.92}{
         \makebox[\textwidth][c]{
\label{tab:rom}
\begin{tabular}{llllll}
\toprule
             & DIP               & PIP              & MCP flex           & MCP abd (extended pose) & MCP abd (flexed pose) \\
             \midrule
Human finger & \, 0--85$^{\circ}$   & 0--105$^{\circ}$ & -45--100$^{\circ}$ & -30--30$^{\circ}$       & -5--5$^{\circ}$~\cite{ishii2020quantitative}       \\
OPH finger   & -9--108$^{\circ}$ & 3--101$^{\circ}$ & -83--110$^{\circ}$ & -41--31$^{\circ}$       & -8--7$^{\circ}$     \\
\bottomrule
\end{tabular}}}
\end{table}

\subsection*{Tendon configurations}

\renewcommand{\thefigure}{S2}
\begin{figure}[ht]
    \centering
    \includegraphics[width=0.75\columnwidth]{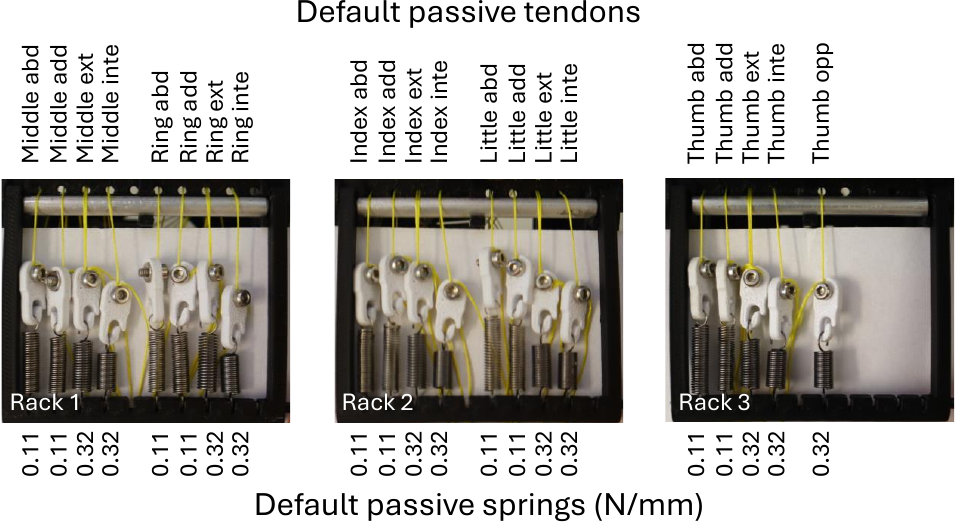}
    \caption{Default passive springs. Passive spring racks are overlayed for compactness and printed tendon clips allow for tendon length reconfiguration.
    }
    \label{fig:s1}
\end{figure}

\renewcommand{\thefigure}{S3}
\begin{figure}[ht]
    \centering
    \includegraphics[width=0.99\columnwidth]{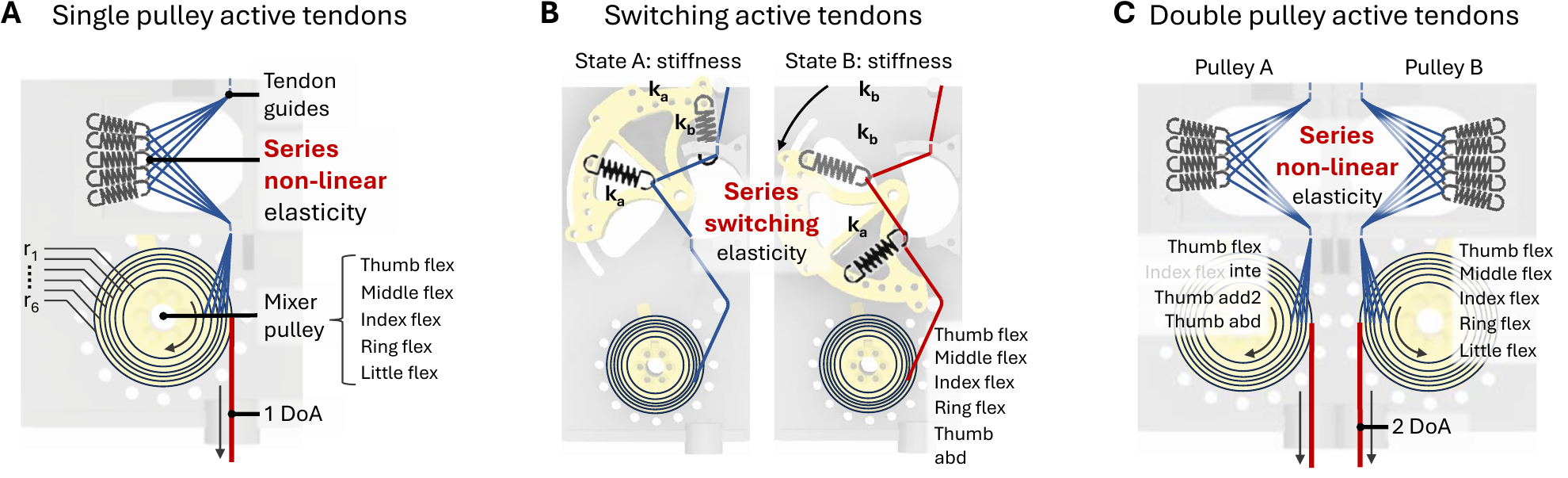}
    \caption{Active spring configurations. (\textbf{A}) Single pulley, (\textbf{B}) switching, (\textbf{C}) double pulley. Series springs are overlayed for vertical compactness. Active tendons, series stiffness, pulley radii, and tendon length can all be varied to alter hand behaviours.
    }
    \label{fig:s2}
\end{figure}

\end{document}